\documentclass[sigconf]{acmart}

\usepackage{algorithm}
\usepackage{algorithmic}
\usepackage{booktabs}
\usepackage{multirow}
\usepackage{geometry}
\usepackage{array}
\usepackage{caption}
\usepackage{booktabs}
\usepackage{newtxmath}
\usepackage{balance}

\AtBeginDocument{%
}

\copyrightyear{2024}
\acmYear{2024}
\setcopyright{rightsretained}
\acmConference[CIKM '24]{Proceedings of the 33rd ACM International Conference on Information and Knowledge Management}{October 21--25, 2024}{Boise, ID, USA}
\acmBooktitle{Proceedings of the 33rd ACM International Conference on Information and Knowledge Management (CIKM '24), October 21--25, 2024, Boise, ID, USA}
\acmDOI{10.1145/3627673.3679748}
\acmISBN{979-8-4007-0436-9/24/10}

\makeatletter
\gdef\@copyrightpermission{
 \begin{minipage}{0.3\columnwidth}
 \href{https://creativecommons.org/licenses/by/4.0/}{\includegraphics[width=0.90\textwidth]{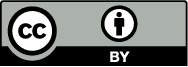}}
 \end{minipage}\hfill
 \begin{minipage}{0.7\columnwidth}
 \href{https://creativecommons.org/licenses/by/4.0/}{This work is licensed under a Creative Commons
Attribution International 4.0 License.}
 \end{minipage}
 \vspace{5pt}
}
\makeatother

\begin{document}

\title{Boosting Certified Robustness for Time Series Classification with Efficient Self-Ensemble}

\author{Chang George Dong}
\orcid{0009-0005-1495-6534}
\email{chang.dong@adelaide.edu.au}
\affiliation{%
  \institution{The University of Adelaide}
  \city{Adelaide}
  \state{SA}
  \country{Australia}
}

\author{Zhengyang David Li}
\orcid{0000-0003-3869-5154}
\email{zhengyang.li01@adelaide.edu.au}
\affiliation{%
  \institution{The University of Adelaide}
  \city{Adelaide}
  \state{SA}
  \country{Australia}
}

\author{Liangwei Nathan Zheng}
\orcid{0009-0007-2793-8110}
\email{liangwei.zheng@adelaide.edu.au}
\affiliation{%
  \institution{The University of Adelaide}
  \city{Adelaide}
  \state{SA}
  \country{Australia}
}

\author{Weitong Chen}
\orcid{0000-0003-1001-7925}
\authornote{Corresponding Author.}
\email{weitong.chen@adelaide.edu.au}
\affiliation{%
  \institution{The University of Adelaide}
  \city{Adelaide}
  \state{SA}
  \country{Australia}
}
\author{Wei Emma Zhang}
\orcid{0000-0002-0406-5974}
\email{wei.e.zhang@adelaide.edu.au}
\affiliation{%
  \institution{ The University of Adelaide}
  \city{Adelaide}
  \state{SA}
  \country{Australia}
}

\renewcommand{\shortauthors}{Chang George Dong, Zhengyang David Li, Liangwei Nathan Zheng, Weitong Chen, \& Wei Emma Zhang}

\begin{abstract}
 Recently, the issue of adversarial robustness in the time series domain has garnered significant attention. However, the available defense mechanisms remain limited, with adversarial training being the predominant approach, though it does not provide theoretical guarantees. Randomized Smoothing has emerged as a standout method due to its ability to certify a provable lower bound on robustness radius under $\ell_p$-ball attacks. Recognizing its success, research in the time series domain has started focusing on these aspects. However, existing research predominantly focuses on time series forecasting, or under the non-$\ell_p$ robustness in statistic feature augmentation for time series classification~(TSC). Our review found that Randomized Smoothing performs modestly in TSC, struggling to provide effective assurances on datasets with poor robustness. Therefore, we propose a self-ensemble method to enhance the lower bound of the probability confidence of predicted labels by reducing the variance of classification margins, thereby certifying a larger radius. This approach also addresses the computational overhead issue of Deep Ensemble~(DE) while remaining competitive and, in some cases, outperforming it in terms of robustness. Both theoretical analysis and experimental results validate the effectiveness of our method, demonstrating superior performance in robustness testing compared to baseline approaches.
\end{abstract}



\maketitle

\section{Introduction}
\noindent\textbf{Background}~~
In recent years, time series data has become increasingly prevalent. As we transit into the era of Industry 4.0, countless sensors generate vast volumes of time series data~\cite{zhang2017eeg,polge2020case,tran2022improving,shen2022death, chen2023death}. Correspondingly, the application of deep neural networks (DNNs) has surged in popularity for time series classification (TSC)\cite{ chen2018eeg,ismail2020inceptiontime,ismail2019deep} as they have achieved remarkable success in various domains especially in computer vision (CV)\cite{krizhevsky2012imagenet,xu2024reliable,zhao2021telecomnet}. However, DNNs exhibit vulnerabilities to minor perturbations in input data, often leading to misclassification and indicating low resistance to external disruption. This issue has attracted significant attention from the research community to the TSC domain. Fawaz et al.~\cite{fawaz2019adversarial} first addressed the weakness of TSC towards adversarial attacks on the UCR dataset. Rathore~\cite{rathore2020untargeted} introduced target and untargeted attacks in TSC. Pialla~\cite{pialla2022smooth} introduced a smooth version of the attack to make it undetectable, which was further improved by Chang~\cite{dong2023swap} who enhanced the stealthiness by leveraging logits information. Ding~\cite{ding2023black} conducted effective black-box attacks on TSC, and Karim~\cite{karim2020adversarial} used adversarial transformation networks to generate adversarial examples.  Other research is progressively expanding the landscape of time series attacks.
\begin{figure*}[ht]
  \centering
  \includegraphics[width=\linewidth]{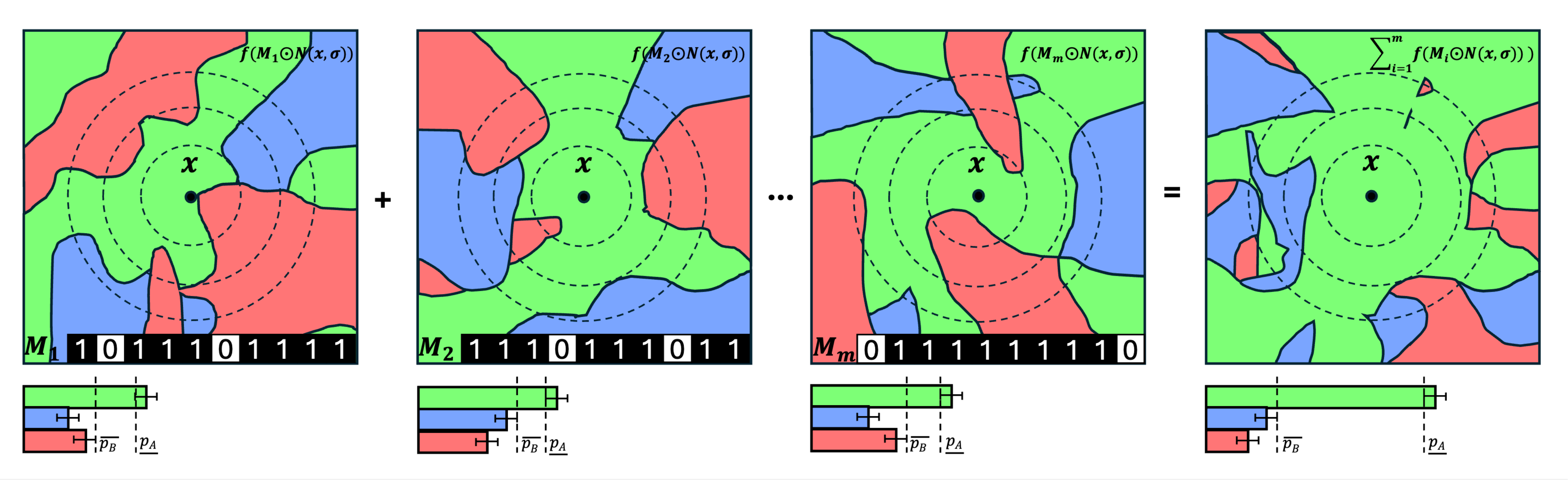}
  \caption{\textbf{Illustration of the Decision Boundary of Base Classifier \( f \) over 3 Classes.} In our method, the decision process is akin to randomly scattering points around \( x \) and counting the number of points that fall into each region to determine the classification confidence. The first three diagrams illustrate the landscapes of three different base classifiers under different fixed masks. After ensembling, the combined base classifier performs better than a single classifier, with the area proportion of \( p_A \)~(where class $A$ is the top one output of the smoothed classifier) significantly increasing, thereby enhancing the certified radius.~(For a detailed proof, please refer to the Theoretical Analysis in Section 4.)
  }
  \label{mainfig}
\end{figure*}
\\[3pt]
In contrast, the development of defenses has been relatively limited. Nevertheless, some works have focused on defending against these attacks. Tariq~\cite{tariq2022towards} proposed using anomaly detection methods to identify adversarial patterns in time series data. Kühne~\cite{kuhne2022defending} combined adversarial training and selective classification to defend against attacks. Abdu~\cite{abdu2022investigating} leveraged random convolutional kernel transformations~(ROCKET) as feature extractors to constitute a robust classifier. Siddiqui~\cite{siddiqui2020benchmarking} reviewed various adversarial training techniques in TSC benchmarking.
\\[3pt]
\noindent\textbf{Challenges}~~
Despite substantial efforts to fortify models against such attacks, challenges persist. Although the above empirical defense methods can enhance robustness in some scenarios, they often fall short as new attacks continuously emerge, rendering existing defenses out of date~\cite{kumari2023trust}. Besides, there is no guarantee above regarding the extent of perturbations a model can withstand. This motivated the study of provable robustness in DNNs to obtain a certified guarantee to resist perturbations. Randomized smoothing, one of the most promising directions in certified defense, was recently proposed by Cohen, et al.~\cite{cohen2019certified}, Li, et al.~\cite{li2019certified}, and Lecuyer, et al.~\cite{lecuyer2019certified} with theoretical guarantees and empirical result. It is simple to transform an arbitrary base classifier $f:\mathbb{R}^d \rightarrow \mathbb{R}^k$ to a smoothed classifier $g$ by training with $i.i.d.$ Gaussian noise. The smoothed classifier $g$ returns the most likely result according to the base classifier under \(x+\epsilon \sim \mathcal{N}(x, \sigma^2 I)\). In simpler terms, instead of predicting a specific point $x$, the model predicts the outcomes for the points surrounding $x$ according to the Gaussian sampling and then combines these predictions through a voting process. This approach helps in making more robust and accurate predictions by considering the uncertainty around each point. 
\\[3pt]
Due to its achievement, in the field of time series, research on robustness has also begun to focus on these aspects. From 2022 to 2023, Yoon, et al.~\cite{yoon2022robust} and Liu, et al.~\cite{liu2023robust} successively applied randomized smoothing techniques to univariate and multivariate time series forecasting tasks, thereby achieving comparatively good robust probabilistic time series forecasting. Additionally, Doppa, et al~\cite{belkhouja2022adversarial} proposed combining randomized smoothing with the statistical features of time series to enhance robustness under non-standard $\ell_p$ balls.
\\[3pt]
However, the aforementioned works~\cite{yoon2022robust,liu2023robust,belkhouja2022adversarial}  did not thoroughly evaluate the specific performance of Randomized Smoothing on various TSC architectures. Our preliminary experiments found that in some cases of adversarial attacks, Randomized Smoothing did not effectively provide defense as expected (See Case Study in \textbf{Section 6.3}). So, how can Randomized Smoothing performance be improved? Currently, there are two main schools of thought on improving Randomized Smoothing: either by carefully designing the noise distribution or by enhancing the performance of the base classifier 
$f$. While training a classifier will be easier compared with the noise distribution design. For example, Salman, et al.~\cite{salman2019provably} combined adversarial training with Randomized Smoothing to enhance performance and Zhai, et al.~\cite{zhai2020macer} directly optimized the target task as a loss. These methods either require lengthy training times or involve complex optimization. While our intuition is more straightforward, improving the performance of the base classifier by variance reduction through aggregation as  Miklós, et al.~\cite{horvath2021boosting}. They found that the Deep Ensemble method~\cite{lakshminarayanan2017simple} can indeed enhance model performance, however, this involves training multiple base classifiers, which can also be problematic in computation cost.
\\[3pt]
\noindent\textbf{Motivation and Contribution}~~
Given the challenges posed by existing methods, we sought to design a simpler approach that does not require multiple training iterations. Initially, we considered methods like MC dropout~\cite{gal2016dropout} and masksemble~\cite{durasov2021masksembles}. MC dropout, being inherently stochastic during prediction, fails to provide theoretical guarantees. Masksemble, on the other hand, involves masking model weights and using a neural network to calculate scores for potential dropout, which complicates training and yields moderate results. Consequently, we propose a self-ensemble method that does not involve training multiple models directly but rather allows a single model to adapt to multiple data distributions as illustrated in Figure \ref{mainfig}. Our contribution can be summarized into 4 folds:

\begin{itemize}
    \item Our method significantly reduces training time by $m$-fold~($m$ is the number of ensembles of the base model). This involves augmenting the training of time series data with masks and then using a few randomly fixed masks for voting classification during prediction. 
    
    \item We provide a detailed theoretical proof of this approach, showing that the self-ensemble method reduces the variance of classification margin in the base classifier, thereby enhancing the confidence lower bound of the top one class, $p_A$. This has also been observed in our experiments. 
    
    \item To claim our findings, we conduct extensive experiments training the base classifier at different noise levels on the UCR benchmark datasets~\cite{dau2019ucr} using three different model architectures~(CNN, RNN, Attention) and evaluated the performance of ensemble methods versus single models. Results show that under CNN and Attention models, self-ensemble significantly outperforms single models, achieving or even exceeding the performance of deep ensembles. 
    
    \item We conduct a case study in Section 6.3 to assess the performance of these methods and benign models under PGD-$\ell_2$ adversarial attacks in different projection radii, finding that self-ensemble can effectively resist adversarial attacks, especially in the model training in the datasets with poor robustness.
\end{itemize}

\section{Related Works}
\subsection{Adversarial Attack}
In TSC, an adversarial attack refers to a malicious attempt to introduce slight perturbations to a time series \( x \in \mathbb{R}^d \) to produce a closely related series \( x' \in \mathbb{R}^d \) with the goal of altering the predicted label. This can be characterized by:
\begin{equation}
\begin{split}
    \text{argmax}\ \{f(x)\} \ne \text{argmax}\ \{f(x')\}, \\
    x' = x + r,\ \text{s.t.} \ ||r||^2 \ll ||x||^2.
\end{split}
\end{equation}
Here, \( f(x) \) represents the predicted probability distribution over the labels for the input \( x \). The perturbation \( r \) is intentionally small in magnitude relative to \( x \) as indicated by their norms.
\\[3pt]
\noindent{Adversarial robustness refers to the ability to resist misclassification caused by adversarial perturbations. 

\subsection{Adversarial Defense}
Adversarial defense can be categorized into empirical and certifiable defense. Early efforts primarily focus on empirical methods such as obfuscated gradients, model distillation, input transformation, adversarial detection, and adversarial training. 
\\[3pt]
\textbf{Some Empirical Defense\ \ }Obfuscated gradients, a defense technique disrupting gradient continuity or hindering gradient acquisition to thwart attacks, were later successfully bypassed by Athalye et al.~\cite{athalye2018obfuscated}, demonstrating their ineffectiveness. Model distillation~\cite{papernot2016distillation} trains a teacher neural network on the original dataset and uses the teacher's class probabilities as soft targets for training a student network, enhancing resilience against adversarial attacks. However, adversaries may adapt their strategies if they know the distillation process. Other methods, like input transformation~\cite{yin2022defending}, aim to convert adversarial samples into clean space using an "auxiliary module" for safe usage. Additionally, the rising popularity of diffusion techniques has led to the use of denoising techniques to remove noise from adversarial samples~\cite{nie2022diffusion}. Anomaly detection also serves as a non-direct defense mechanism in adversarial settings~\cite{tariq2022towards}.
\\[3pt]
\textbf{Adversarial Traning\ \ }Empirically, the most practical method is adversarial training. This method involves finding the perturbation from the maximum point within the loss surface using Projected Gradient Descent~(PGD),  then minimizing it during the normal training process~\cite{madry2017towards}. Alternating between these steps can smooth the loss surface near the input point, leading to a smaller Lipschitz constant. This makes it harder for attackers to find adversarial patterns near the input. The PGD adversarial training process can be described as follows:}
\begin{equation}
\delta = \arg\max_{\delta, \|\delta\|_p \leq \epsilon} L(f(x + \delta), y),
\end{equation}
where \( L \) is the loss function. The model is then trained to minimize this maximum loss:
\begin{equation}
\min_\theta \mathbb{E}_{(x,y) \sim \mathcal{D}} \left[ \max_{\|\delta\|_p \leq \epsilon} L(f_\theta(x + \delta), y) \right].
\end{equation}
Here, \( \theta \) represents the model parameters and \( \mathcal{D} \) is the training data distribution respectively. However, adversarial training is time-consuming and lacks a theoretical guarantee.
\\[3pt]
\noindent{\textbf{Certified Defense\ \ }
In comparison with empirical defense, the certified defense can provide a theoretical guarantee rather than experimental findings. It can be defined as the following work flow. Given a classifier \( f : \mathcal{X} \rightarrow \mathcal{Y}
\in \{1, \ldots, k\} \), if \( f \) can correctly classify all samples to the same label within an \( l_p \)-ball of radius \( r \) centered at \( x \), then the classifier is robust around \( x \) against \( l_p \) attacks with radius \( r \). This indicates that the classifier maintains its predictive consistency for all inputs within this ball.}

\section{Methodology}

\subsection{Preliminary}

\noindent\textbf{Randomized Smoothing}~~Randomized Smoothing has recently been proposed~\cite{cohen2019certified, li2019certified} as its power to resist adversarial attacks in a tight theoretical certified space. It is simple to transform an arbitrary base classifier $f:\mathbb{R}^d \rightarrow \mathbb{R}^k$ to a smoothed classifier $g$ by training with $i.i.d.$ Gaussian noise. The smoothed classifier $g$ returns the most likely result according to the base classifier under \(x+\epsilon \sim \mathcal{N}(x, \sigma^2 I)\), i.e.

\begin{equation}
    g(x) = \arg\max_{c \in \mathcal{Y}} \mathbb{P}_{\delta \sim \mathcal{N}(0, \sigma^2 I)}(f(x + \delta) = c),
\end{equation}
where the noise level \(\sigma\) is a hyperparameter to trade off the accuracy and robustness. In other words, $g$ will return the class c whose decision region \{\(x' \in \mathbb{R}^d: f(x') = c\)\} has the largest probability measure under the distribution of $\mathcal{N}(x, \sigma^2 I)$. Cohen, et al.~\cite{cohen2019certified} recently proved a tight robustness guarantee in $\ell_2$  norms for smoothing with Gaussian noise. Meanwhile, Lecuyer, et al.~\cite{lecuyer2019certified} and Li, et al.~\cite{li2019certified} also demonstrated that the smoothed classifier $g$ will consistently classify within a certified radius around the input $x$ under $l_2$ norm considerations, based on Differential Privacy~(DP) and Rényi divergence~\cite{van2014renyi} respectively~\cite{kumari2023trust}. In this paper, our theoretical analysis chooses divergence-based randomized smoothing from Li, et al.~\cite{li2019certified}. Details are provided in the \textbf{THEORETICAL ANALYSIS} section, \textbf{Theorem 4.1}~\cite{li2019certified}.
\\[3pt]
\noindent\textbf{Deep Ensemble for Randomized Smoothing}~~
Ensemble methods, classical in statistical machine learning, enhance predictive performance by aggregating multiple models to reduce their inter-covariance. Introduced in 2016~\cite{lakshminarayanan2017simple}, Deep Ensembles were designed to reduce uncertainty in neural network predictions. This approach has been adapted for Randomized Smoothing~(RS)~\cite{horvath2021boosting,qin2021dynamic,liu2020enhancing}, which narrows the prediction distribution range, effectively raising the lower bound of predictions. In this context, ensemble methods with Randomized Smoothing involve employing multiple base classifiers trained with identical architectures but different random seeds. Which was defined as:
\begin{equation}
    \hat{f}(x) = \frac{1}{m} \sum_{l=1}^{m} f_l(x),
\end{equation}
where $f_l(x)$ are the logits of the outputs from classifiers, and each classifier $f_l$ is trained independently using a unique seed. This configuration ensures diversity among the classifiers, which reduces variance to increase the lower confidence bound of the true majority class probability $p_A$. However, Deep Ensembles require extensive retraining, leading to resource inefficiency. To address this, we propose the following method. 

\subsection{Proposed Method}

\textbf{Random Mask Training}~~We propose a training algorithm that incorporates randomized masking based on a normal randomized smoothing procedure, as seen in \textbf{Algorithm 1}. Our intuition is to help models adapt to different types of masks during training, ensuring that the prediction confidence shows small differences among various masks~(Details about how to implement masks are described in \textbf{5.3 Algorithm Setup}). This method can avoid multiple training of deep ensemble, which greatly reduces the computational overhead. In the inference stage, we will leverage the diversity of these masks to obtain the self-ensemble, and the efficacy is competitive and even better in some cases compared with the deep-ensemble.

\begin{algorithm}
  \caption{Training Base Model \(f\) with Noise Under Mask}
  \begin{algorithmic}[1]
    \REQUIRE Initialize classifier \( f_{\theta_0} \) over \(\{1, \dots, k\}\); Probability \( p \) for mask; Training set  \( D = \{(x_i, y_i)\}_{i=1}^n \); Sequence Length \(T\); A standard deviation \( \sigma > 0\);
    \FOR{$epoch = 1$ to $n$}
    \STATE \# Generate random binomial masks \( M \) for each time series \( x \) in the dataset, with a probability \( p \) of being 1 and sequence length \( T \).
    \STATE \( M \leftarrow \text{BINOMIALMASK}(T, p) \)
    \STATE \# Train with random Gaussian noise under the mask
    \STATE \( \theta_{i+1} \leftarrow \theta_i - \nabla L(f_{\theta_i}(M \odot{N(x,\sigma^2 I)}), y) \)
    \ENDFOR
  \end{algorithmic}
\end{algorithm}

\noindent\textbf{Self-ensemble Certification}~To certify the robustness of the smoothed classifier \( g \) around \( x \), we designed the process as shown in \textbf{Algorithm 2}. In this framework, a set of fixed masks is randomly generated and kept consistent for all inputs. During the ensemble inference stage, a single sample undergoes \( m \times n \) trials, where \( m \) and \( n \) are the total numbers of different masks and noises, respectively. For each noised input \( x \), predictions based on these \( m \) masks are aggregated to produce the final output. The class resulting from that iteration increments its count by one. This process will iteratively run $n$ times before the final prediction is determined by hard voting across the \( n \) noised inputs. We then calculate the confidence intervals based on the prediction counts for the top one and top two classes $A$ and $B$ using a multinomial distribution, taking the lower bound of \( \underline{p_A} \) and the upper bound of \( \overline{p_B} \) as conservative estimates of \( p_A \) and \( p_B \). Based on the prediction, we obtain the lower bound of the certified radius \( L \). This means that for all \( |x' - x| < L \), the prediction will remain consistent.

\noindent{The reason why we need to fix the masks is to ensure that the model will return identical results for the same input. This is a fundamental requirement of \textbf{Theorem 4.1}~\cite{li2019certified}. Compared to the training process, masks play a different role in the certification stage. The same model with multiple masks can have a similar effect as a deep ensemble, significantly reducing the variance of \( p_A \) and \( p_B \) and increasing the gaps between \( \underline{p_A} \) and \( \overline{p_B} \). This results in an increased certified radius, making the model more robust.}

\begin{algorithm}
  \caption{Certify Robust Classifier \(g\) Around \(x\)}
  \begin{algorithmic}[1]
    \REQUIRE A pre-trained classifier \( f \) over \( \{1, \ldots, k\} \); Probability \( p \) for mask; An input time series \( x \) with sequence Length \(T\); A standard deviation \( \sigma > 0 \); Number of certify iterations \( n \); The confidence level \(\beta\); Parameter series \(\alpha >1\). Ensemble Size \(m\).
    \FOR{\( i = 1 \) to \( m \)}
    \STATE \( M_i \leftarrow \text{BINOMIALMASK}(T, p, seed(i)) \)
    \ENDFOR
    \FOR{\( i = 1 \) to \( n \)}
    \STATE \( Class_i \leftarrow \mathop{\operatorname*{arg\,max}}_{k}\frac{1}{m} \sum_{i=1}^m f(M_i \odot{N(x,\sigma^2 I)}) \)

    \STATE \(Counts[\hat{Class_i}] += 1\)
    \ENDFOR
    \STATE Prediction: \(\hat{g(x)} \leftarrow \mathop{\operatorname*{arg\,max}}_{k}Counts \)
    \STATE Top two: \(\underline{p_A},\overline{p_B} \leftarrow MultiNominalCI(Counts, \beta)\)
    \STATE Conservative estimate: \(\hat{p_A},\hat{p_B}  \leftarrow \underline{p_A},\overline{p_B} \)
    \STATE Let: \(M_p(x_1,...,x_n) \leftarrow~(\frac{1}{n}\sum_{i=1}^n x_i^p)^{1/p}\)
    \STATE \(L = sup_{\substack{\alpha > 1}}(-\frac{2\sigma^2}{\alpha}\ \log{(1-2M_1(p_A,p_B) + 2M_{1-\alpha}(p_A,p_B)))^{1/2}}\)
    \RETURN Prediction \(\hat{g(x)}\), Certified Radius Lower Bound \(L \)
  \end{algorithmic}
\end{algorithm}

\section{Theoretical Analysis}
To prove our claimed findings, we provide a detailed theoretical analysis. Starting with \textbf{Theorem 4.1} proved by Li et al.~\cite{li2019certified}, a tight lower bound for the certified radius of an arbitrary classifier \( g \) based on Rényi divergence is described as follows:
\\[3pt]
\noindent{\textbf{Theorem 4.1\ \ }~(From Li, et al~\cite{li2019certified}). \textit{Suppose $\mathbf{x} \in \mathcal{X}$, and a potential adversarial example $\mathbf{x}' \in \mathcal{X}$, such that $\|\mathbf{x} - \mathbf{x}'\|_2 \leq L$. Given a $k$-classifier $f : \mathcal{X}\in\mathbb{R}^d \rightarrow p\in\mathbb{R}^k$, let $f(\mathcal{N}(\mathbf{x}, \sigma^2 I)) \sim~(p_1, \ldots, p_k)$ and $f( \mathcal{N}(\mathbf{x'}, \sigma^2 I)) \sim~(p'_1, \ldots, p'_k)$, and $M_p(x_1,...,x_n) \leftarrow~(\frac{1}{n}\sum_{i=1}^n x_i^p)^{1/p}.$ If the following condition is satisfied, with $p_{A}$ and $p_{B}$ being the first and second largest probabilities in $\{p_i\}$:}}
\begin{equation}
\sup_{\alpha > 1} \left( -\frac{2\sigma^2}{\alpha} \log\left(1 - 2M_1(p_{A}, p_{B}) + 2M_{1-\alpha}(p_{A}, p_{B})\right) \right) \geq L^2
,\label{Theorem 4.1 eq1 }\end{equation}
\textit{then \(\mathop{\operatorname*{arg\,max}}_{i} p_i = \mathop{\operatorname*{arg\,max}}_{j} p'_j.\)}
\\[3pt]
To increase the lower bound of \( L \), it is intuitive to increase the noise level \( \sigma \), however, the model may struggle with the noised input, leading to a drop in accuracy. In turn, increase the gaps between \( p_A \) and \( p_B \) can also raise \( L \). Here, we propose the self-ensemble method, which can reduce the variance of predictions \( p_A \) and \( p_B \) and enlarge their gaps. The variance and expectation after self-ensemble can be described as follows:
\\[3pt] 
\textbf{Lemma 4.2\ \ }\textit{
Let \( f \) be a pre-trained classifier with masks and noises over \(\mathbb{R}^d \to \mathbb{R}^k\). Let \( M_i \) be a set of \( m \) independent binomial masks applied to the input \( \mathbf{x} \) with probability \( p \rightarrow 1 \)  of each element being 1. Assume that the input noise \(\epsilon \sim \mathcal{N}(0, \sigma^2 I)\). The $m$-ensemble average model output is given by:}

\begin{equation}
  {\overline{y}} = \frac{1}{m} \sum_{i=1}^m f(M_i \odot~(\mathbf{x} + \epsilon)).
\end{equation}
\textit{The expectation and variance of the ensemble average output are:}
\begin{equation}
  \mathbb{E}[\overline{y}] = \mathbb{E}[f(\mathbf{x})] = \mathbf{c},
\end{equation}
\textit{where $\mathbf{c}\in \mathbb{R}^k$ is the expectation output of a fixed clean $x$ over randomness in the training process. }

\begin{equation}
    \begin{split}
    \text{Var}(\overline{y}) & = \frac{1}{m}~(1+\zeta_c~(m-1)) \Sigma_c \\
    & + \frac{1}{m}~(1+\zeta_p~(m-1)) \sigma^2 \mathbb{E}[J_f(M \odot \mathbf{x}) J_f(M \odot \mathbf{x})^T]),
    \end{split}
\end{equation}
\textit{where \(\Sigma_c\) is the covariance matrix representing the variability across different training processes, characterizing the randomness. \(J_f(M \odot \mathbf{x})\) is the Jacobian matrix of \( f \). \(\zeta_c\) is the correlation coefficient for the masked output in the clean part, and \(\zeta_p\) is the correlation coefficient for the masked Jacobian in the perturbed part.}
\\[3pt]
\noindent{\textbf{Proof 4.2\ \ } Let \( y_i = f(M_i \odot~(\mathbf{x} + \epsilon)) \) be the output for each mask. Since \( M_i \) and \(\epsilon\) are independent, and \(\epsilon\) has mean 0, we have:}
\begin{equation}
    \mathbb{E}[y_i] = \mathbb{E}[f(M_i \odot~(\mathbf{x} + \epsilon))] = \mathbb{E}[f(M_i \odot \mathbf{x})].
\end{equation}
During the training process, we observed the network can achieve the same performance with and without the mask, especially when \( p \) is close to 1, the mask has minimal influence on the prediction, thus:
\begin{equation}
 \mathbb{E}[f(\mathbf{x})] = \mathbb{E}[f(M \odot \mathbf{x})] = \mathbf{c}.
\end{equation}
Averaging over multiple independent \( M_i \):
\begin{equation}
 \mathbb{E}[\overline{y}] = \mathbb{E}\left[\frac{1}{m} \sum_{i=1}^m y_i\right] = \frac{1}{m} \sum_{i=1}^m \mathbb{E}[y_i] = \mathbb{E}[f(\mathbf{x})] = \mathbf{c}.
\end{equation}
Now, we compute the variance of the ensemble average output. According to the Taylor Expansion, \( y_i = f(M_i \odot \mathbf{x}) + M_i \odot J_f(\mathbf{x}) \epsilon \).
For the variance of a single \( y_i \):
\begin{equation}
\text{Var}(y_i) = \Sigma_c + \sigma^2 J_f(M_i \odot \mathbf{x}) J_f(M_i \odot \mathbf{x})^T.
\end{equation}
For the ensemble average output \( \overline{y} \):
\begin{equation}
     \text{Var}(\overline{y}) = \text{Var}\left(\frac{1}{m} \sum_{i=1}^m y_i\right).
\end{equation}
Assuming the correlation between \( y_i \) and \( y_j \) is parameterized by coefficients \( \zeta_c \) and \( \zeta_p \), we have:
\begin{equation}
    \text{Var}(\overline{y}) = \frac{1}{m^2} \left( \sum_{i=1}^m \text{Var}(y_i) + \sum_{i \neq j} \text{Cov}(y_i, y_j) \right),
\end{equation}
where \textit{$\text{Cov}(y_i, y_j) = \zeta_c \Sigma_c + \zeta_p \sigma^2 J_f(M_i \odot \mathbf{x}) J_f(M_i \odot \mathbf{x})^T$}. Therefore:
\begin{equation}
    \begin{split}
    \text{Var}(\overline{y}) & = \frac{1}{m}~(1+\zeta_c~(m-1)) \Sigma_c \\
    &+ \frac{1}{m}~(1+\zeta_p~(m-1)) \sigma^2 \mathbb{E}[J_f(M \odot \mathbf{x}) J_f(M \odot \mathbf{x})^T]).
    \end{split}
\end{equation}
Thus, the Lemma is proved.
\\[3pt] 
\textbf{Variance Reduction\ \ } When \( m = 1 \) and the mask is all ones, the variance degenerates to:
\begin{equation}
    \text{Var}(y) = \Sigma_c + \sigma^2 J_f(\mathbf{x}) J_f(\mathbf{x})^T.
\end{equation}
Instead, after mask ensemble, the two parts of the variance are reduced to \( \frac{1}{m}~(1 + \zeta_c~(m-1)) \) and \( \frac{1}{m}~(1 + \zeta_p~(m-1)) \) times of the original, respectively. When the correlation coefficients are less than 1, these values will always be less than 1 and approach the correlation coefficients \(\zeta_c\) and \(\zeta_p\) respectively as \( m \) increases.
Additionally, the reduction in variance after masking also comes from the shrinkage of the Jacobian matrix. Due to the implementation of the mask, the Jacobian matrix should be:
\begin{equation}
    \mathbb{E}[J_f(M \odot \mathbf{x}) J_f(M \odot \mathbf{x})^T] = p^2 J_f(\mathbf{x}) J_f(\mathbf{x})^T.
\end{equation}
Here we assume that the Jacobian matrix is similar in both masked and unmasked training settings. Although in some cases they might not be exactly the same, the difference is relatively small compared to the factor \( p^2 \). Therefore, the variance after applying the mask ensemble is:
\begin{equation}
    \begin{split}
    \text{Var}(\overline{y}) & \approx \frac{1}{m}~(1 + \zeta_c~(m-1)) \Sigma_c \\
     & + \frac{1}{m}~(1 + \zeta_p~(m-1)) \sigma^2 p^2 J_f(\mathbf{x}) J_f(\mathbf{x})^T.
    \end{split}
\end{equation}
As the number of ensembles $m$ approaches infinity, the variance can be further simplified to:
\begin{equation}
\lim_{m\rightarrow \infty}\text{Var}(\overline{y}) \approx \zeta_c \Sigma_c + \zeta_p  p^2 \sigma^2 J_f(\mathbf{x}) J_f(\mathbf{x})^T.
\end{equation}
This result demonstrates that the variance of the ensemble model is reduced due to both the correlation coefficients and the shrinkage of the Jacobian matrix resulting from the masking process.

\begin{figure}[ht]
  \centering
  \includegraphics[width=\linewidth]{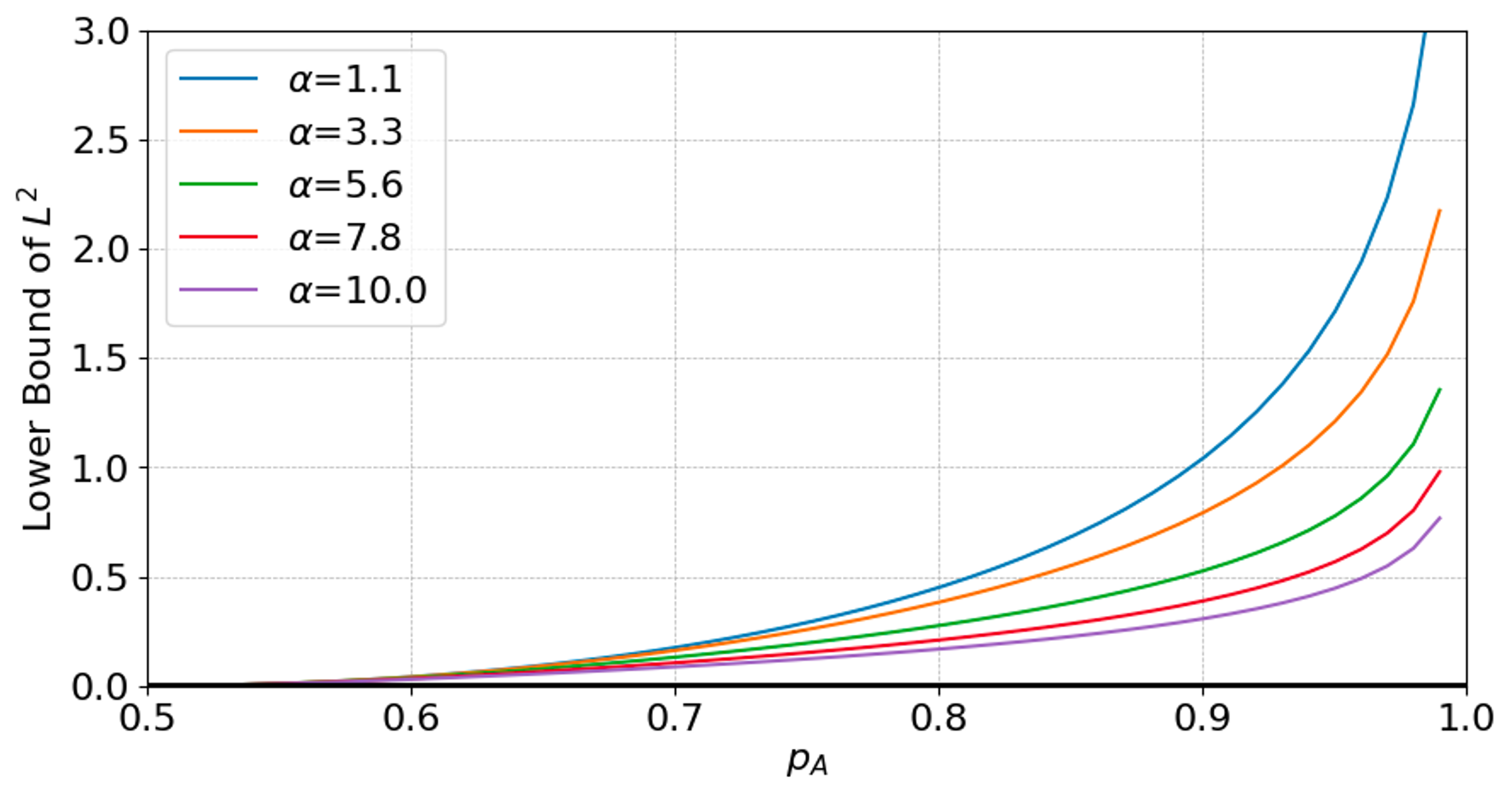}
  \caption{$L^2$ vs. $p_A$ for different values of $\alpha$.  Certified radius can be prompted with increasing the $p_A$ in different $\alpha$.}
  \label{Theoreom2}
\end{figure}

\begin{figure}[ht]
  \centering
  \includegraphics[width=\linewidth]{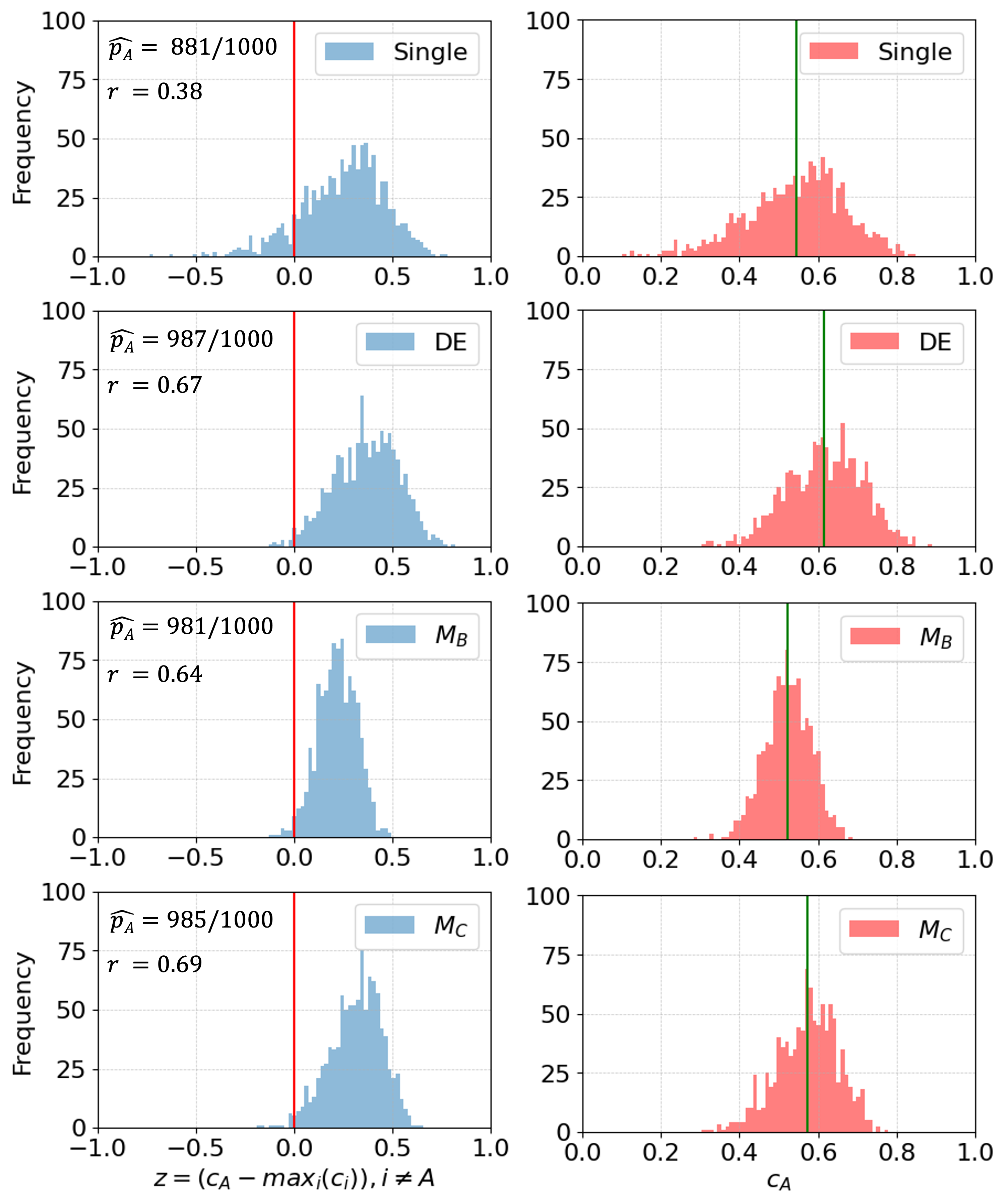}
  \caption{Effect of variance reduction on: Left) classification margins $z$. As the variance decreases through different ensembling methods, the probability $P(z > 0)$ increases, resulting in a higher certified radius. The red line represents the decision boundary at $z = 0$. Right) Corresponding prediction distribution of $c_A$. It exhibits a similar normal distribution, and a tightened distribution can be clearly observed in the ensemble. Additionally, the green line represents the mean. Except for DE, the expectation of the self-ensemble method is closer to the theoretical assumption compared to the Single. The subfigures show the distribution of these values for different ensembling methods: Single, DE, $M_B$, and $M_C$.~(From ChlorineConcentration Dataset, $\sigma = 0.4$, 1000 noise samples)}
  \label{Theoreom1}
\end{figure}

\noindent\textbf{Theorem 4.3\ \ }
\textit{Let $f$ be a pre-trained classifier over $x \in \mathbb{R}^d \to y \in \mathbb{R}^k$. Given an input $x$, let $x' = x + \epsilon$ where $\epsilon \sim \mathcal{N}(0, \sigma^2 I)$. Let $c_A$ be the score for class $A$ and $c_i$ be the score for any other class $i \neq A$. The classification margins are defined as $z_i = c_A - c_i,  \forall i \neq A$. Then, the probability $P_A = P(z_i > 0  , \forall  i \neq A)$ increases as the variance of $\overline{y}$ decreases.}

\noindent{\textbf{Proof 4.3\ \ } According to the local linear assumption, the output of the classifier is perturbed as $f(x') \sim f(x) + A \cdot \epsilon$, where $A$ is an auxiliary matrix. Thus, $f(x') \sim \mathcal{N}(f(x), A\sigma^2A^T)$. According to the variance reduction of $\overline{y}$ from \textbf{Lemma 4.2}, we can know that the auxiliary matrix represents the reduction adjustment of the variance to each class. Thus we know that $A$ should be a matrix with all elements < $1$. Let $c_A \sim \mathcal{N}(\mu_A, \sigma_A^2)$ and $c_i \sim \mathcal{N}(\mu_i, \sigma_i^2)$ for $i \neq A$. The classification margins $z_i$ are:}
\begin{equation}
    z_i = c_A - c_i \sim \mathcal{N}(\mu_{z_i}, \sigma_{z_i}^2).
\end{equation}
Here, $\mu_{z_i} = \mu_A - \mu_i$ and $\sigma_{z_i}^2 = \sigma_A^2 + \sigma_i^2$. Since $z_i \sim \mathcal{N}(\mu_{z_i}, \sigma_{z_i}^2)$, the probability $P(z_i > 0)$ is:
\begin{equation}
    P(z_i > 0) = \Phi\left( \frac{\mu_{z_i}}{\sigma_{z_i}} \right),
\end{equation}
where $\Phi$ is the CDF of the standard normal distribution.
As the variance $\sigma_{z_i}^2$ decreases, $\frac{\mu_{z_i}}{\sigma_{z_i}}$ increases because $\mu_{z_i}$ is fixed. Then, we have:
\begin{equation}
    \Phi\left( \frac{\mu_{z_i}}{\sigma_{z_i,\text{new}}} \right) > \Phi\left( \frac{\mu_{z_i}}{\sigma_{z_i}} \right),
\end{equation}
where $\sigma_{z_i,\text{new}} < \sigma_{z_i}$. Therefore, $P(z_i > 0)$ increases as $\sigma_{z_i}^2$ decreases. Using the union bound $\forall i \neq A$, we have:
\begin{equation}
    P_A = P(z_i > 0,\ \forall  i \neq A) \geq \prod_{i \neq A} P(z_i > 0).
\end{equation}
As each $P(z_i > 0)$ increases with decreasing $\sigma_{z_i}^2$, the overall probability $P_A$ increases. Thus, as the variance $\sigma_{z_i}^2$ decreases, the success probability $P_A$ increases. Thus, as $P_A$ increases, we can simply set $P_B = 1 - P_A$ to make a conservative estimation. Then, the lower bound $L$ in Theorem 4.1~(Li et al.) will increase. As depicted in Figure \ref{Theoreom2}, when we fix the noise level $\sigma$, higher $p_A$ can guarantee a larger radius.
\\[3pt]

We further validated our theoretical findings through experiments. Figure 3 illustrates a sample case from our benchmark dataset, ChlorineConcentration Dataset, with a standard deviation~($\sigma$) of 0.4 and 1000 noise samples. The distribution of the classification margins $z$ of the ensemble method exhibits fewer values distributed less than 0, indicating a higher probability of being classified as class A~(the top 1 prediction of the smoothed classifier $g$), thus certifying a larger radius. Additionally, the right four distributions in Figure 4 depict the distribution of $c_A(x)$, revealing a noticeable variance reduction of the $z$ distribution in the self-ensemble method compared with the Single one. Notably, the mean values~(represented by the green line) of $M_B$ and $M_C$ closely align with the Single method, supporting our assumption that the mask should have minimal influence on the expectation value.

\section{Experiments}
\subsection{Experimental Setup}

\noindent The whole project was developed using PyTorch, and conducted on a server equipped with Nvidia RTX 4090 GPUs, 64 GB RAM, and an AMD EPYC 7320 processor. Table \ref{training time} gives the training time of each method over different datasets in InceptionTime architecture (other models show a similar trend), which claimed our contribution that self-ensemble takes less time for the training process compared with Deep Ensemble (5 models ensemble).     \href{https://github.com/Chang-George-Dong/Boosting-Certificate-Robustness-for-Time-Series-Classification-with-Efficient-Self-Ensemble.git}{[GitHub]}

\subsection{Datasets}
To evaluate the proposed method and compare it with the baseline, we conduct our experiment in diverse univariate time series benchmark datasets~\cite{dau2019ucr}. 
Table \ref{description of dataset} gives a detailed description of these datasets. For the case study, we selected the ChlorineConcentration and Cricket datasets for our demonstration, as ChlorineConcentration is the most susceptible to adversarial attacks among these datasets, while CricketX is more robust. Thereby to provide a clearer illustration of the efficacy of all methods.

\begin{table}[t!]
\centering
\caption{Training time of each method over different datasets}
\label{training time}
\begin{tabular}{c|cccc}
\toprule
Dataset & Single & DE & $M_B$ & $M_C$ \\
\midrule
ChlorineConcentration & 276.5 & 1338.1 & 274.5 & 275.2 \\
SyntheticControl      & 25.1  & 139.9  & 25.3  & 24.6  \\
CBF                   & 46.1  & 232.0  & 45.8  & 45.3  \\
CricketX              & 110.2 & 530.3  & 111.0 & 110.2 \\
CricketY              & 105.0 & 525.3  & 104.7 & 103.9 \\
CricketZ              & 105.0 & 530.6  & 105.0 & 104.5 \\
\bottomrule
\end{tabular}
\end{table}

\begin{table}[h]
\centering
\small
\caption{Detailed description of the datasets}
\label{description of dataset}
\begin{tabular}{c|ccccc}
\toprule
Name & Type & Train & Test & Class & Length \\
\midrule
ChlorineConcentration & Sensor & 467 & 3840 & 3 & 166 \\
SyntheticControl & Simulated & 300 & 300 & 6 & 60 \\
CBF & Simulated & 30 & 900 & 3 & 128 \\
CricketX & Motion & 390 & 390 & 12 & 300 \\
CricketY & Motion & 390 & 390 & 12 & 300 \\
CricketZ & Motion & 390 & 390 & 12 & 300 \\
\bottomrule
\end{tabular}
\end{table}

\begin{table*}[t!]
\centering
\small
\renewcommand{\arraystretch}{0.9}
\caption{Comprehensive performance comparison of various smoothing approaches~(Single, DE, $M_B$ and $M_C$) under different architectural settings across multiple datasets with varying $\sigma$. The top value represents the Average Certified Radius~(ACR), while the bottom value with $\%$ denotes the corresponding accuracy.
}
\label{table1}
\begin{tabular}{c|c|c|ccc|c|ccc|c|ccc}
\hline
 &  & \multicolumn{4}{c|}{InceptionTime} & \multicolumn{4}{c|}{LSTM-FCN} & \multicolumn{4}{c}{MACNN} \\

\cline{3-14}
       Dataset&    $\sigma$    & Single & DE & M$_B$ & M$_C$ & Single & DE & M$_B$ & M$_C$&Single & DE & M$_B$ & M$_C$\\
\hline
\multirow{10}{*}{ChlorineConcentration} 
&0.1	&0.144	&\textbf{0.162}	&0.156	&0.146	&\textbf{0.166}	&0.126	&0.136	&0.116	&0.155	&0.166	&\textbf{0.18}	&0.157\\

&	&82.1\%	&82.0\%	&76.4\%	&81.0\%	&69.3\%	&76.3\%	&64.6\%	&69.6\%	&80.0\%	&79.4\%	&67.3\%	&69.9\%\\

&0.2	&0.188	&0.201	&0.246	&\textbf{0.257}	&0.147	&0.182	&0.238	&\textbf{0.254}	&0.191	&0.239	&\textbf{0.307}	&0.292 \\

&	&74.6\%	&75.4\%	&66.8\%	&70.6\%	&59.1\%	&69.8\%	&64.1\%	&63.4\%	&70.3\%	&68.4\%	&61.8\%	&62.6\%\\

&0.4	&0.243	&0.361	&0.532	&\textbf{0.540}	&0.489	&0.505	&\textbf{0.562}	&0.559	&0.481	&0.567	&0.547	&\textbf{0.687}\\

&    &59.5\%	&61.7\%	&58.5\%	&60.4\%	&59.1\%	&58.8\%	&58.3\%	&58.5\%	&58.8\%	&59.1\%	&57.0\%	&57.5\%\\

&0.8	&0.816	&1.151	&\textbf{1.549}	&1.156	&1.217	&1.291	&\textbf{1.342}	&1.342	&\textbf{1.509}	&1.382	&1.394	&1.473\\

&	&56.1\%	&56.2\%	&55.9\%	&56.2\%	&56.4\%	&56.2\%	&56.4\%	&56.8\%	&56.2\%	&56.4\%	&56.6\%	&56.2\%\\

&1.6	&3.228	&3.654	&3.265	&\textbf{3.752}	&3.300	&\textbf{3.576}	&3.521	&3.481	&\textbf{3.329}	&2.86	&3.094	&3.069\\

&	&54.8\%	&54.7\%	&54.9\%	&54.5\%	&54.3\%	&53.7\%	&53.9\%	&54.5\%	&54.7\%	&55.1\%	&54.8\%	&54.8\%\\

\hline
\multirow{10}{*}{SyntheticControl} 
&0.1	&\textbf{0.247}	&0.247	&0.243	&0.244	&\textbf{0.245}	&0.245	&0.235	&0.241	&0.244	&\textbf{0.245}	&0.244	&0.244\\

&&99.7\%	&99.7\%	&91.3\%	&96.3\%	&98.7\%	&98.3\%	&94.3\%	&97.3\%	&100.0\%	&99.7\%	&97.7\%	&98.0\%\\

&0.2	&\textbf{0.490}	&0.489	&0.456	&0.453	&0.483	&\textbf{0.486}	&0.453	&0.465	&0.483	&\textbf{0.486}	&0.465	&0.467\\

&&99.7\%	&99.7\%	&95.3\%	&95.3\%	&98.3\%	&98.0\%	&94.7\%	&96.7\%	&99.7\%	&99.7\%	&98.0\%	&97.7\%\\

&0.4	&0.902	&\textbf{0.912}	&0.808	&0.776	&0.873	&\textbf{0.885}	&0.800	&0.785	&0.902	&\textbf{0.910}	&0.796	&0.775\\

&&99.7\%	&99.7\%	&84.0\%	&91.0\%	&99.3\%	&98.7\%	&95.7\%	&95.3\%	&99.7\%	&99.7\%	&90.0\%	&86.7\%\\

&0.8	&1.245	&1.274	&1.069	&\textbf{1.431}	&1.204	&\textbf{1.226}	&1.060	&1.020	&1.237	&\textbf{1.265}	&0.971	&1.056\\

&&100.0\%	&100.0\%	&83.0\%	&79.3\%	&99.3\%	&99.3\%	&90.3\%	&92.0\%	&99.7\%	&99.7\%	&91.3\%	&85.7\%\\

&1.6	&1.182	&1.196	&\textbf{1.469}	&1.315	&1.099	&\textbf{1.123}	&0.947	&1.110	&1.181	&\textbf{1.193}	&0.883	&0.720\\

&&98.7\%	&98.3\%	&45.0\%	&65.7\%	&98.3\%	&98.3\%	&85.7\%	&79.3\%	&98.7\%	&98.7\%	&77.7\%	&74.7\%\\

\hline
\multirow{10}{*}{CBF} 
&0.1	&\textbf{0.248}	&0.248	&0.248	&0.248	&0.244	&\textbf{0.245}	&0.242	&0.244	&0.246	&0.247	&\textbf{0.248}	&0.247\\

&&99.9\%	&99.9\%	&100.0\%	&100.0\%	&97.7\%	&99.1\%	&97.2\%	&98.1\%	&99.6\%	&99.3\%	&99.8\%	&99.3\%\\

&0.2	&0.494	&0.494	&\textbf{0.495}	&0.495	&0.477	&\textbf{0.481}	&0.476	&0.475	&\textbf{0.493}	&0.493	&0.493	&0.492\\

&&100.0\%	&99.9\%	&100.0\%	&100.0\%	&96.9\%	&98.8\%	&96.7\%	&97.4\%	&99.2\%	&99.3\%	&99.8\%	&99.8\%\\

&0.4	&0.979	&\textbf{0.981}	&0.978	&0.979	&0.905	&\textbf{0.912}	&0.901	&0.899	&\textbf{0.975}	&0.969	&0.972	&0.97\\

&&100.0\%	&100.0\%	&99.9\%	&100.0\%	&95.4\%	&97.8\%	&96.8\%	&96.7\%	&100.0\%	&99.8\%	&99.7\%	&100.0\%\\

&0.8	&1.576	&1.627	&\textbf{1.661}	&1.639	&\textbf{1.552}	&1.551	&1.547	&1.544	&1.618	&1.667	&1.647	&\textbf{1.699}\\

&&99.7\%	&99.6\%	&99.4\%	&100.0\%	&91.0\%	&93.9\%	&93.7\%	&93.3\%	&99.7\%	&99.6\%	&99.4\%	&99.4\%\\

&1.6	&\textbf{2.051}	&1.980	&1.956	&1.967	&1.709	&\textbf{1.752}	&1.705	&1.709	&1.960	&\textbf{1.993}	&1.955	&1.918\\

&&93.1\%	&95.6\%	&90.9\%	&95.9\%	&88.4\%	&90.1\%	&93.3\%	&92.3\%	&94.9\%	&97.4\%	&91.2\%	&92.7\%\\

\hline
\multirow{10}{*}{CricketX} 
&0.1	&0.230	&0.236	&\textbf{0.240}	&0.232	&\textbf{0.237}	&0.236	&0.230	&0.234	&0.219	&0.227	&\textbf{0.233}	&0.232\\

&&82.3\%	&85.4\%	&83.6\%	&84.4\%	&72.8\%	&74.6\%	&68.5\%	&70.5\%	&82.3\%	&82.1\%	&79.2\%	&83.3\%\\

&0.2	&0.436	&0.454	&\textbf{0.462}	&0.456	&0.441	&\textbf{0.454}	&0.449	&0.451	&0.440	&0.436	&0.445	&\textbf{0.451}\\

&&83.3\%	&84.6\%	&82.3\%	&82.3\%	&73.8\%	&74.9\%	&70.3\%	&68.7\%	&82.3\%	&83.6\%	&80.5\%	&83.3\%\\

&0.4	&0.814	&0.862	&0.862	&\textbf{0.874}	&0.798	&\textbf{0.823}	&0.817	&0.816	&0.825	&0.836	&0.862	&\textbf{0.885}\\

&&83.1\%	&84.6\%	&84.4\%	&81.5\%	&72.6\%	&74.1\%	&68.7\%	&67.2\%	&84.6\%	&85.4\%	&77.9\%	&81.8\%\\

&0.8	&1.463	&1.520	&1.537	&\textbf{1.575}	&1.360	&\textbf{1.464}	&1.339	&1.342	&1.508	&\textbf{1.560}	&1.513	&1.513\\

&&82.3\%	&83.3\%	&80.5\%	&80.0\%	&68.5\%	&69.5\%	&67.2\%	&66.7\%	&81.8\%	&81.0\%	&79.7\%	&79.5\%\\

&1.6	&2.078	&2.182	&\textbf{2.189}	&2.182	&1.722	&\textbf{1.928}	&1.805	&1.782	&2.309	&2.292	&\textbf{2.315}	&2.240\\

&&78.5\%	&78.7\%	&75.4\%	&74.4\%	&65.6\%	&66.2\%	&63.1\%	&62.8\%	&69.7\%	&73.6\%	&61.3\%	&62.8\%\\

\hline
\multirow{10}{*}{CricketY} 
&0.1	&0.226	&\textbf{0.236}	&0.227	&0.231	&0.234	&0.233	&\textbf{0.235}	&0.227	&0.222	&0.222	&\textbf{0.232}	&0.224\\

&&82.6\%	&85.9\%	&81.3\%	&81.5\%	&74.1\%	&76.2\%	&70.5\%	&72.8\%	&82.1\%	&83.3\%	&79.0\%	&84.9\%\\

&0.2	&0.428	&0.443	&\textbf{0.445}	&0.431	&\textbf{0.439}	&0.439	&0.439	&0.429	&0.414	&0.422	&0.425	&\textbf{0.433}\\

&&83.1\%	&84.4\%	&82.6\%	&81.0\%	&73.6\%	&74.9\%	&69.2\%	&69.5\%	&82.3\%	&83.1\%	&80.0\%	&81.3\%\\

&0.4	&0.776	&\textbf{0.806}	&0.795	&0.786	&0.742	&\textbf{0.814}	&0.773	&0.761	&0.771	&\textbf{0.79}	&0.737	&0.757\\

&&83.6\%	&82.8\%	&80.5\%	&80.5\%	&74.1\%	&70.8\%	&68.7\%	&66.2\%	&81.5\%	&81.8\%	&79.7\%	&81.0\%\\

&0.8	&1.281	&\textbf{1.342}	&1.313	&1.317	&1.181	&\textbf{1.297}	&1.151	&1.147	&1.297	&\textbf{1.332}	&1.312	&1.286\\

&&81.0\%	&81.3\%	&78.5\%	&78.2\%	&68.5\%	&67.9\%	&66.9\%	&64.9\%	&80.5\%	&81.0\%	&75.9\%	&78.5\%\\

&1.6	&1.562	&1.699	&1.734	&\textbf{1.746}	&1.399	&\textbf{1.609}	&1.408	&1.365	&1.619	&1.777	&1.699	&\textbf{1.965}\\

&&77.4\%	&76.9\%	&74.1\%	&72.6\%	&63.3\%	&64.4\%	&63.1\%	&62.3\%	&75.4\%	&74.4\%	&64.9\%	&62.1\%\\

\hline
\multirow{10}{*}{CricketZ} 
&0.1	&0.237	&\textbf{0.240}	&0.239	&0.237	&0.228	&0.232	&\textbf{0.238}	&0.231	&0.229	&0.234	&0.227	&\textbf{0.236}\\

&&83.1\%	&86.2\%	&82.1\%	&86.4\%	&78.5\%	&77.9\%	&73.1\%	&76.2\%	&83.6\%	&84.4\%	&79.5\%	&82.1\%\\

&0.2	&0.454	&\textbf{0.46}	&0.457	&0.460	&0.435	&\textbf{0.456}	&0.455	&0.451	&0.441	&0.447	&0.443	&\textbf{0.456}\\

&&82.3\%	&85.4\%	&84.1\%	&86.4\%	&77.7\%	&76.7\%	&71.8\%	&74.4\%	&85.4\%	&85.9\%	&80.3\%	&84.4\%\\

&0.4	&0.825	&0.836	&0.840	&\textbf{0.855}	&0.815	&\textbf{0.851}	&0.827	&0.793	&0.830	&\textbf{0.867}	&0.828	&0.850\\

&&83.1\%	&84.6\%	&83.8\%	&83.8\%	&74.9\%	&74.9\%	&70.8\%	&73.1\%	&84.1\%	&82.1\%	&81.5\%	&84.4\%\\

&0.8	&1.433	&1.498	&1.512	&\textbf{1.522}	&1.328	&\textbf{1.434}	&1.293	&1.331	&1.538	&1.540	&1.524	&\textbf{1.550}\\

&&82.6\%	&83.6\%	&83.1\%	&83.1\%	&71.5\%	&71.5\%	&70.3\%	&70.0\%	&80.5\%	&81.5\%	&79.7\%	&80.8\%\\

&1.6	&1.983	&\textbf{2.118}	&2.086	&2.112	&1.605	&\textbf{1.829}	&1.704	&1.688	&2.034	&2.118	&\textbf{2.165}	&2.142\\

&&79.2\%	&79.2\%	&78.5\%	&78.2\%	&67.2\%	&68.2\%	&65.6\%	&66.4\%	&77.4\%	&79.0\%	&75.6\%	&78.7\%\\

\hline
\end{tabular}
\end{table*}

\subsection{Algorithm Setup}
We implemented three different DNN architectures: \textbf{InceptionTime}~(CNN)~\cite{ismail2020inceptiontime}, \textbf{LSTM-FCN}~(RNN)~\cite{karim2017lstm}, and \textbf{MACNN}~(Attention)~\cite{chen2021multi}. All models were trained with noise level \(\sigma\) of \textbf{0, 0.1, 0.2, 0.4, 0.8} and \textbf{1.6} for 1000 epochs to achieve a smoothed version. We used two kinds of masking settings: binomial masking and continuous masking. In binomial masking~(\(M_B\)), for each timestamp, there is a probability $p$ to keep the original value; otherwise, it is set to \textbf{0}. In continuous binomial masking~(\(M_{C}\)), there are several continuous mask-lets, combing to the total masking length fixed to $p$ times the sequence length, and the maximum mask segment length set to half of $p$ times the sequence length. Unless specified otherwise, all $p$ were set to \textbf{0.9}. During the certifying phase, we found that 1000 noise draws were sufficient, and set confidence level \(\beta =\) \textbf{0.001}. The certified noise level  \(\sigma\) used the same settings as the training phase. For adversarial attacks, we used the PGD-$\ell_2$  attacks from Madary~\cite{madry2017towards} with different epsilons, as listed in the result tables for evaluation.

\subsection{Evaluation Metrics}
We evaluate two key metrics on these models:~(i) the certified accuracy at predetermined radius $r$ and~(ii) the average certified radius~(ACR). In the case study, we evaluate the adversarial robustness using the Attack Success Rate which is the proportion of misclassified samples by adversarial perturbation.
\section{Result}

\begin{figure*}[ht]
  \centering
\includegraphics[width=\linewidth]{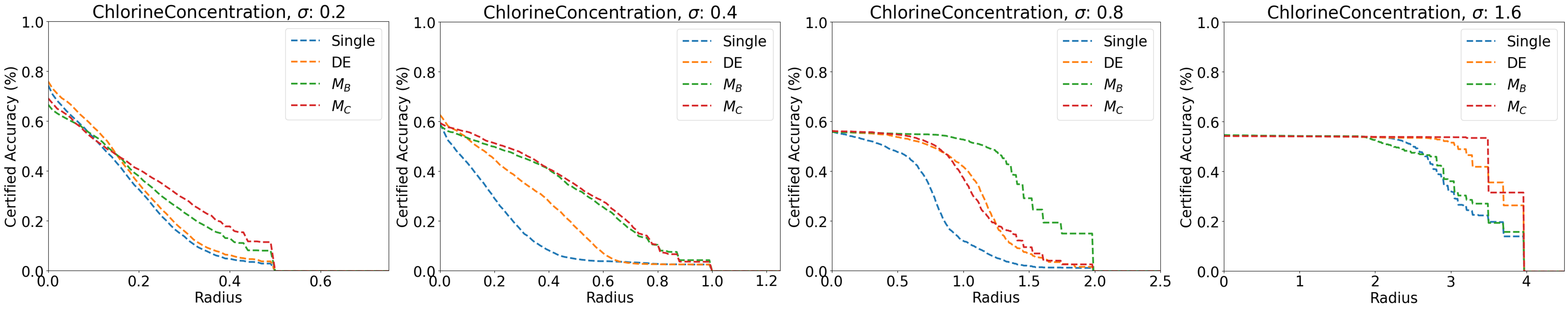}
  \caption{Certified Accuracy vs. radius in different $\sigma$ on ChlorineConcentration.~($\sigma$ is $0.2,\ 0.4,\ 0.8\ $and$\ 1.6$ from left to right)}
  \label{Certified Accuracy}
\end{figure*}

\subsection{Main Results}
\textbf{Comprehensive Performance Comparison\ \ }In Table \ref{table1}, we compare our self-ensemble method using two kinds of masks, $M_B$ and $M_C$, Deep Ensembles~(DE), and the Single model across three different types of networks, trained using five different noise levels $\sigma$. The results show a consistent trend: as the noise level $\sigma$ increases, the Averaged Certified Radius~(ACR) increases, while the accuracy decreases. In almost every case, ensemble models perform better compared to their individual counterparts. This implies masking is effective in a wide range of datasets. 
\\[3pt]
Notably, the self-ensemble methods using $M_B$ and $M_C$ both perform well on InceptionTime and MACNN networks, often competing with and even outperforming the DE method, while requiring only 1/5 of the training time~(all ensemble methods in this table are 5-model ensembles). However, for the LSTM-FCN architecture, $M_B$ and $M_C$ perform worse than the single model. This is primarily because sequential models are highly sensitive to missing values; each input in a sequence model is critical, and a single missing value can easily skew the model's predictions.  In contrast, convolutional models are more robust to missing values, especially when the receptive field is large. Convolutions can learn from neighboring values and effectively handle local missing patterns. Missing values in convolutions can be mitigated by the surrounding non-missing values, reducing the bias introduced by the missing data. 
Therefore, it is safer to use masks in convolutional and attention-based networks.
\\[3pt]
\textbf{Certifed Accuracy over Radius\ \ }
In comparison with the stationary performance shown in Table \ref{table1}, Figure \ref{Certified Accuracy} illustrates the performance under increasing perturbations, which better reflects real-world scenarios with adversarial noise. We observe that the self-ensemble method is resilient to a wider range of perturbations~(indicated by the radius in the figure) while maintaining accuracy, followed by the DE and Single models. However, as the radius increases, the accuracy of all models tends to deteriorate,  as explained by Madry et al., "Robustness May Be at Odds with Accuracy"~\cite{tsipras2018robustness}. Nevertheless, the degradation is slower in the ensemble methods compared to the single model, highlighting the benefits of ensemble approaches in maintaining performance under adversarial conditions.

\subsection{Ablation Study} 

\begin{figure}[htpb]
  \centering
\includegraphics[width=\linewidth]{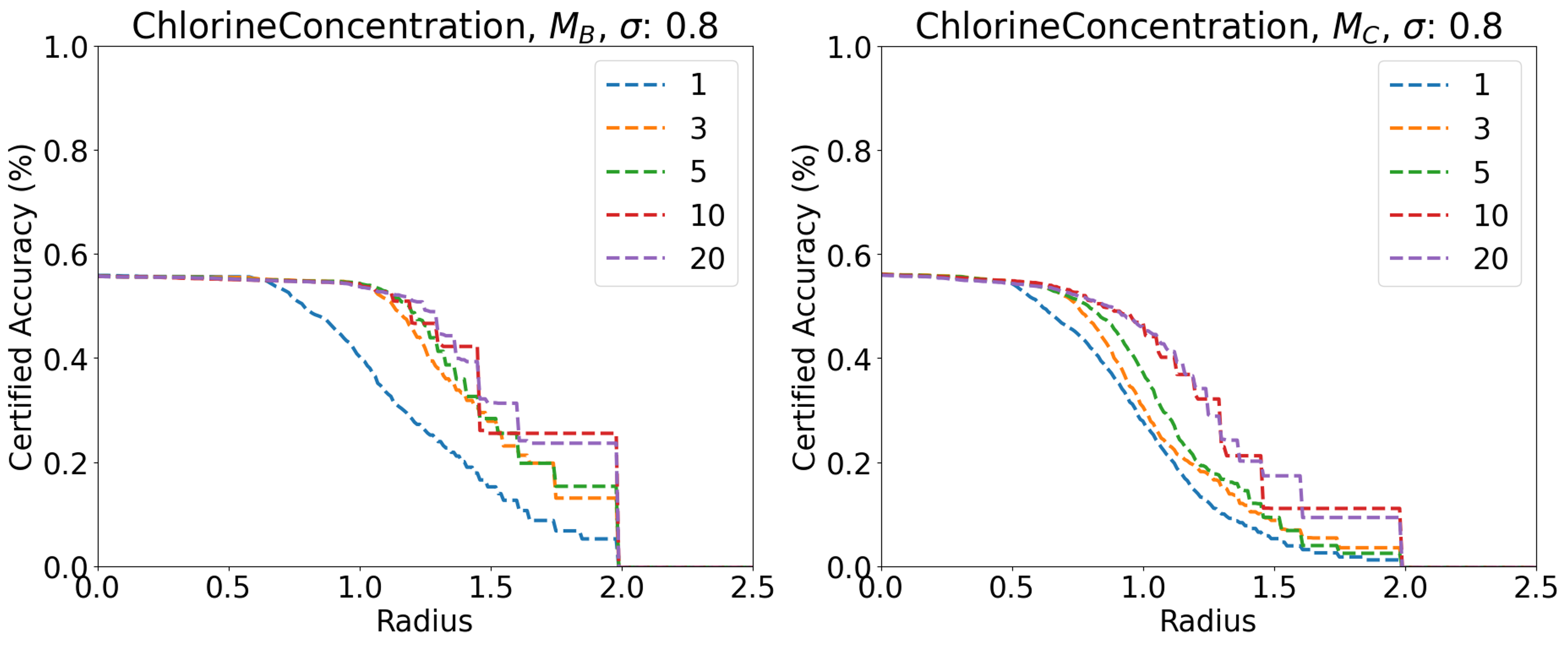}
  \caption{Certified accuracy vs. radius for different ensemble sizes on ChlorineConcentration.(Left: $M_B$, Right: $M_C$)}
  \label{Ensemble Size}
\end{figure}

\noindent\textbf{Ensemble Size\ \ } Figure \ref{Ensemble Size} shows the certified accuracy vs. radius for different ensemble sizes. Both methods demonstrate that increasing the ensemble size improves certified robustness. However, the improvement tends to saturate when the ensemble size reaches 10 for $M_C$ and 5 for $M_B$, while requiring double and quadruple the inference time, respectively. Therefore, we choose an ensemble size of 5 as it provides a good balance between performance and computational efficiency.

\begin{figure}[htpb]
  \centering
\includegraphics[width=\linewidth]{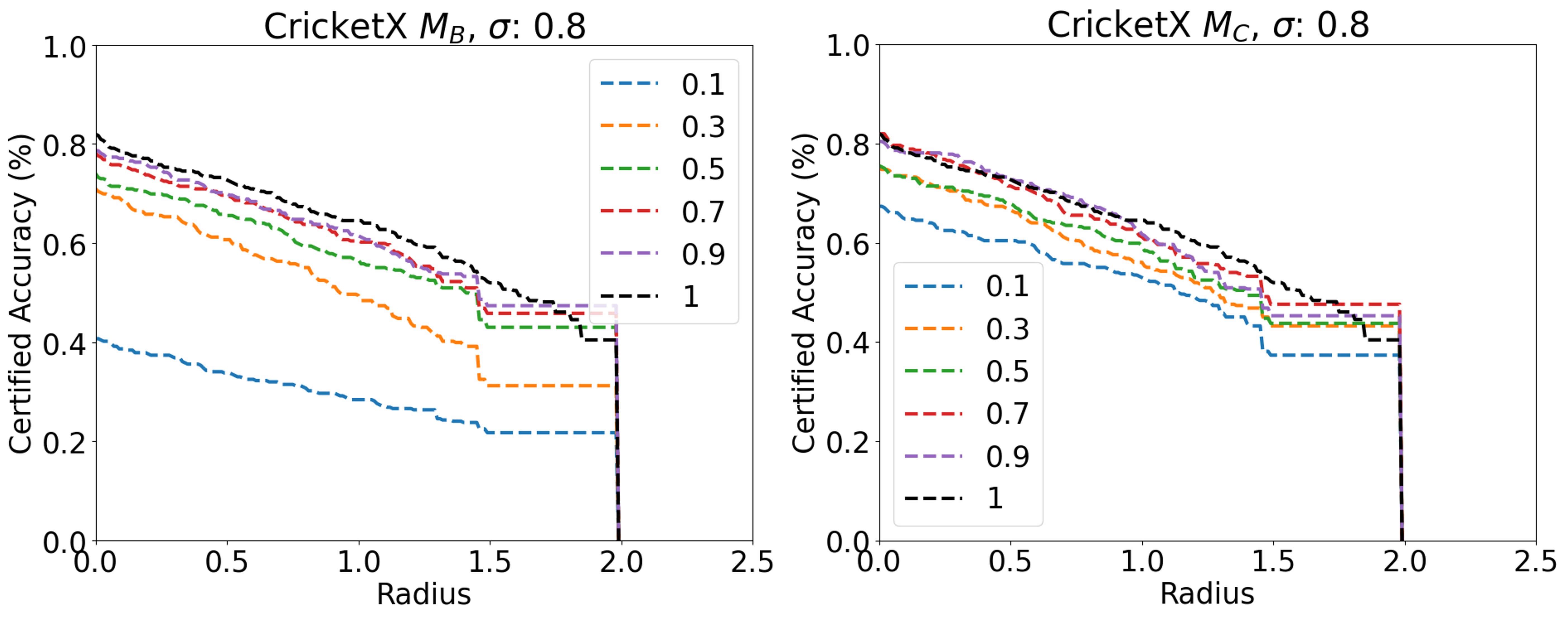}
  \caption{Certified accuracy vs. radius for different keep ratio on CricketX.(Left: $M_B$, Right: $M_C$)}
  \label{Mask Ratio}
\end{figure}

\begin{table}[h!]
\centering
\setlength{\abovecaptionskip}{10pt}
\setlength{\belowcaptionskip}{10pt}
\setlength{\textfloatsep}{10pt}

\caption{Attack Success Rate~(ASR) vs. perturbation level $\epsilon$ under PGD-$\ell_2$ attack in benign models and various randomized smoothing classifiers~($\sigma = 0.4$) on ChlorineConcentration~(CC) and CricketX datasets}
\label{table2}
\small
\begin{tabular}{c|c|ccccc}
\toprule
Dataset & $\epsilon$ & Benign & Single & DE & $M_B$ & $M_C$ \\
\midrule
\multirow{4}{*}{CC} & 0.25 &   0.98 &   0.55 & 0.35 &   \textbf{0.24} &    0.26 \\
& 0.50 &   0.99 &   0.79 & 0.54 &   \textbf{0.47} &    0.50 \\
& 0.75 &   1.00 &   0.93 & 0.74 &   \textbf{0.69} &    0.71 \\
& 1.00 &   1.00 &   0.96 & 0.93 &   \textbf{0.84} &    0.87 \\
\midrule
\multirow{4}{*}{CricketX} & 0.25 &   0.22 &   0.12 & \textbf{0.08} &   0.10 &    0.09 \\
& 0.50 &   0.43 &   0.18 & \textbf{0.14} &   0.18 &    0.20 \\
& 0.75 &   0.57 &   0.24 & \textbf{0.18} &   0.24 &    0.31 \\
& 1.00 &   0.67 &   0.32 & \textbf{0.25} &   0.33 &    0.39 \\
\bottomrule
\end{tabular}
\end{table}

\noindent{\textbf{Keep Ratio\ \ }
Figure \ref{Mask Ratio} shows the influence of the keep ratio on the accuracy-radius curve. It is evident that a high keep ratio~(masking to 1) $p$ in both cases ensures better-certified accuracy over the radius and is close to the performance of the non-mask method. This is highly aligned with our theoretical assumptions, a high keep ratio is very close to no masking. The same trend clarifies that a higher keep ratio is practically important to align with the theoretical analysis. Additionally, models tend to be more sensitive to a low keep ratio of $M_B$ compared to $M_C$, as $M_C$ can retain more continuous patterns that help convolution correctly activate. This implies that a segment is sufficient for CNN recognition, and it may be possible to alleviate the dimensionality curse~\cite{anderson2022certified} in Randomized Smoothing by distilling to the input size.}

\subsection{Case Study: Adversarial Robustness}

To validate our method's performance on datasets with insufficient robustness compared to the Single model, we also conducted adversarial attacks using PGD-$\ell_2$ against the Benign models, single smoothed, and various ensemble smoothed classifiers ($\sigma = 0.4$). Table \ref{table2} illustrates the Attack Success Rate (ASR) versus the perturbation level $\epsilon$ under PGD-$\ell_2$ attack on both the ChlorineConcentration (CC) and CricketX datasets. The results clearly demonstrate that the ensemble approach is significantly more effective in mitigating attacks compared to the Single model. Notably, on the ChlorineConcentration dataset, $M_B$ achieves a remarkable reduction in attack success rate from 0.98 to nearly 0.24, whereas the Single model remains vulnerable with a $55\%$ ASR. As mentioned before, the Single model did not demonstrate strong robustness on this dataset. In contrast, on the more robust CricketX dataset, all four methods perform similarly, with DE being slightly more effective. These findings underscore the robustness of self-ensemble methods in defending against adversarial attacks, highlighting their superiority over individual models.

\section{Conclusion}

In this paper, we utilized self-ensemble to decrease the variance of classification margins, thereby boosting the lower bound of top-1 prediction confidence and certifying a larger radius to enhance model robustness. Theoretical proofs and experimental findings substantiate the effectiveness of our approach, leveraging ensemble techniques while mitigating computational overhead. Our case study illustrates superior robustness compared to baseline methods against adversarial cases.
\\[6pt]
\textbf{Future work\ \ } Despite its efficacy, we observed suboptimal performance in RNN architectures, and it may attributed to their sensitivity to missing values. To address this, we aim to train sequential models resistant to the random missing value in the future. Additionally, in the certification process, the use of random fixed masks may lead to unstable results. Hence, we plan to focus on mask design to enhance the stability of this algorithm.

\section{Acknowledgments}
We thank all the creators and providers of the UCR time series benchmark datasets and the valuable suggestions of Dr. Eamonn Keogh for our work.
This research work is supported by \textit{Australian Research Council Linkage Project} (LP230200821),
\textit{Australian Research Council Discovery Projects} (DP240103070 ),
\textit{Australian Research Council ARC Early Career Industry Fellowship} (IE230100119), 
\textit{Australian Research Council ARC Early Career Industry Fellowship} (IE240100275), and \textit{University of Adelaide, Sustainability FAME Strategy Internal Grant 2023}.
\newpage
\bibliographystyle{unsrt}
\balance
\bibliography{sample-base}

\begin{thebibliography}{10}

\bibitem{zhang2017eeg}
Dalin Zhang, Lina Yao, Xiang Zhang, Sen Wang, Weitong Chen, and Robert Boots.
\newblock Eeg-based intention recognition from spatio-temporal representations
  via cascade and parallel convolutional recurrent neural networks.
\newblock {\em arXiv preprint arXiv:1708.06578}, pages 1--8, 2017.

\bibitem{polge2020case}
Julien Polge, J{\'e}r{\'e}my Robert, and Yves Le~Traon.
\newblock A case driven study of the use of time series classification for
  flexibility in industry 4.0.
\newblock {\em Sensors}, 20(24):7273, 2020.

\bibitem{tran2022improving}
Khai~Phan Tran, Weitong Chen, and Miao Xu.
\newblock Improving traffic load prediction with multi-modality: A case study
  of brisbane.
\newblock In {\em Australasian Joint Conference on Artificial Intelligence},
  pages 254--266. Springer, 2022.

\bibitem{shen2022death}
Shaofei Shen, Miao Xu, Lin Yue, Robert Boots, and Weitong Chen.
\newblock Death comes but why: An interpretable illness severity predictions in
  icu.
\newblock In {\em Asia-Pacific Web (APWeb) and Web-Age Information Management
  (WAIM) Joint International Conference on Web and Big Data}, pages 60--75.
  Springer, 2022.

\bibitem{chen2023death}
Weitong Chen, Wei~Emma Zhang, and Lin Yue.
\newblock Death comes but why: A multi-task memory-fused prediction for
  accurate and explainable illness severity in icus.
\newblock {\em World Wide Web}, 26(6):4025--4045, 2023.

\bibitem{chen2018eeg}
Weitong Chen, Sen Wang, Xiang Zhang, Lina Yao, Lin Yue, Buyue Qian, and Xue Li.
\newblock Eeg-based motion intention recognition via multi-task rnns.
\newblock In {\em Proceedings of the 2018 SIAM International Conference on Data
  Mining}, pages 279--287. SIAM, 2018.

\bibitem{ismail2020inceptiontime}
Hassan Ismail~Fawaz, Benjamin Lucas, Germain Forestier, Charlotte Pelletier,
  Daniel~F Schmidt, Jonathan Weber, Geoffrey~I Webb, Lhassane Idoumghar,
  Pierre-Alain Muller, and Fran{\c{c}}ois Petitjean.
\newblock Inceptiontime: Finding alexnet for time series classification.
\newblock {\em Data Mining and Knowledge Discovery}, 34(6):1936--1962, 2020.

\bibitem{ismail2019deep}
Hassan Ismail~Fawaz, Germain Forestier, Jonathan Weber, Lhassane Idoumghar, and
  Pierre-Alain Muller.
\newblock Deep learning for time series classification: a review.
\newblock {\em Data mining and knowledge discovery}, 33(4):917--963, 2019.

\bibitem{krizhevsky2012imagenet}
Alex Krizhevsky, Ilya Sutskever, and Geoffrey~E Hinton.
\newblock Imagenet classification with deep convolutional neural networks.
\newblock {\em Advances in neural information processing systems}, 25, 2012.

\bibitem{xu2024reliable}
Cai Xu, Jiajun Si, Ziyu Guan, Wei Zhao, Yue Wu, and Xiyue Gao.
\newblock Reliable conflictive multi-view learning.
\newblock In {\em Proceedings of the AAAI Conference on Artificial
  Intelligence}, volume~38, pages 16129--16137, 2024.

\bibitem{zhao2021telecomnet}
Wei Zhao, Ziyu Xu, Cai~andGuan, Xunlian Wu, Wanqing Zhao, Qiguang Miao, Xiaofei
  He, and Quan Wang.
\newblock Telecomnet: Tag-based weakly-supervised modally cooperative hashing
  network for image retrieval.
\newblock {\em IEEE Transactions on Pattern Analysis and Machine Intelligence},
  44(11):7940--7954, 2021.

\bibitem{fawaz2019adversarial}
Hassan~Ismail Fawaz, Germain Forestier, Jonathan Weber, Lhassane Idoumghar, and
  Pierre-Alain Muller.
\newblock Adversarial attacks on deep neural networks for time series
  classification.
\newblock In {\em 2019 International Joint Conference on Neural Networks
  (IJCNN)}, pages 1--8. IEEE, 2019.

\bibitem{rathore2020untargeted}
Pradeep Rathore, Arghya Basak, Sri~Harsha Nistala, and Venkataramana Runkana.
\newblock Untargeted, targeted and universal adversarial attacks and defenses
  on time series.
\newblock In {\em 2020 international joint conference on neural networks
  (IJCNN)}, pages 1--8. IEEE, 2020.

\bibitem{pialla2022smooth}
Gautier Pialla, Hassan~Ismail Fawaz, Maxime Devanne, Jonathan Weber, Lhassane
  Idoumghar, Pierre-Alain Muller, Christoph Bergmeir, Daniel Schmidt, Geoffrey
  Webb, and Germain Forestier.
\newblock Smooth perturbations for time series adversarial attacks.
\newblock In {\em Pacific-Asia Conference on Knowledge Discovery and Data
  Mining}, pages 485--496. Springer, 2022.

\bibitem{dong2023swap}
Chang~George Dong, Liangwei~Nathan Zheng, Weitong Chen, Wei~Emma Zhang, and Lin
  Yue.
\newblock Swap: Exploiting second-ranked logits for adversarial attacks on time
  series.
\newblock In {\em 2023 IEEE International Conference on Knowledge Graph
  (ICKG)}, pages 117--125. IEEE, 2023.

\bibitem{ding2023black}
Daizong Ding, Mi~Zhang, Fuli Feng, Yuanmin Huang, Erling Jiang, and Min Yang.
\newblock Black-box adversarial attack on time series classification.
\newblock In {\em Proceedings of the AAAI Conference on Artificial
  Intelligence}, volume~37, pages 7358--7368, 2023.

\bibitem{karim2020adversarial}
Fazle Karim, Somshubra Majumdar, and Houshang Darabi.
\newblock Adversarial attacks on time series.
\newblock {\em IEEE transactions on pattern analysis and machine intelligence},
  43(10):3309--3320, 2020.

\bibitem{tariq2022towards}
Shahroz Tariq, Binh~M Le, and Simon~S Woo.
\newblock Towards an awareness of time series anomaly detection models'
  adversarial vulnerability.
\newblock In {\em Proceedings of the 31st ACM International Conference on
  Information \& Knowledge Management}, pages 3534--3544, 2022.

\bibitem{kuhne2022defending}
Joana K{\"u}hne and Clemens G{\"u}hmann.
\newblock Defending against adversarial attacks on time-series with selective
  classification.
\newblock In {\em 2022 Prognostics and Health Management Conference (PHM-2022
  London)}, pages 169--175. IEEE, 2022.

\bibitem{abdu2022investigating}
Mubarak~G Abdu-Aguye, Walid Gomaa, Yasushi Makihara, and Yasushi Yagi.
\newblock Investigating strategies towards adversarially robust time series
  classification.
\newblock {\em Pattern Recognition Letters}, 156:104--111, 2022.

\bibitem{siddiqui2020benchmarking}
Shoaib~Ahmed Siddiqui, Andreas Dengel, and Sheraz Ahmed.
\newblock Benchmarking adversarial attacks and defenses for time-series data.
\newblock In {\em International Conference on Neural Information Processing},
  pages 544--554. Springer, 2020.

\bibitem{kumari2023trust}
Anupriya Kumari, Devansh Bhardwaj, Sukrit Jindal, and Sarthak Gupta.
\newblock Trust, but verify: A survey of randomized smoothing techniques.
\newblock {\em arXiv preprint arXiv:2312.12608}, 2023.

\bibitem{cohen2019certified}
Jeremy Cohen, Elan Rosenfeld, and Zico Kolter.
\newblock Certified adversarial robustness via randomized smoothing.
\newblock In {\em international conference on machine learning}, pages
  1310--1320. PMLR, 2019.

\bibitem{li2019certified}
Bai Li, Changyou Chen, Wenlin Wang, and Lawrence Carin.
\newblock Certified adversarial robustness with additive noise.
\newblock {\em Advances in neural information processing systems}, 32, 2019.

\bibitem{lecuyer2019certified}
Mathias Lecuyer, Vaggelis Atlidakis, Roxana Geambasu, Daniel Hsu, and Suman
  Jana.
\newblock Certified robustness to adversarial examples with differential
  privacy.
\newblock In {\em 2019 IEEE symposium on security and privacy (SP)}, pages
  656--672. IEEE, 2019.

\bibitem{yoon2022robust}
TaeHo Yoon, Youngsuk Park, Ernest~K Ryu, and Yuyang Wang.
\newblock Robust probabilistic time series forecasting.
\newblock In {\em International Conference on Artificial Intelligence and
  Statistics}, pages 1336--1358. PMLR, 2022.

\bibitem{liu2023robust}
Linbo Liu, Youngsuk Park, Trong~Nghia Hoang, Hilaf Hasson, and Jun Huan.
\newblock Robust multivariate time-series forecasting: Adversarial attacks and
  defense mechanisms.
\newblock In {\em Proceedings of the International Conference on Learning
  Representations (ICLR)}, 2023.

\bibitem{belkhouja2022adversarial}
Taha Belkhouja and Janardhan~Rao Doppa.
\newblock Adversarial framework with certified robustness for time-series
  domain via statistical features.
\newblock {\em Journal of Artificial Intelligence Research}, 73:1435--1471,
  2022.

\bibitem{salman2019provably}
Hadi Salman, Jerry Li, Ilya Razenshteyn, Pengchuan Zhang, Huan Zhang, Sebastien
  Bubeck, and Greg Yang.
\newblock Provably robust deep learning via adversarially trained smoothed
  classifiers.
\newblock {\em Advances in neural information processing systems}, 32, 2019.

\bibitem{zhai2020macer}
Runtian Zhai, Chen Dan, Di~He, Huan Zhang, Boqing Gong, Pradeep Ravikumar,
  Cho-Jui Hsieh, and Liwei Wang.
\newblock Macer: Attack-free and scalable robust training via maximizing
  certified radius.
\newblock {\em arXiv preprint arXiv:2001.02378}, 2020.

\bibitem{horvath2021boosting}
Mikl{\'o}s~Z Horv{\'a}th, Mark~Niklas M{\"u}ller, Marc Fischer, and Martin
  Vechev.
\newblock Boosting randomized smoothing with variance reduced classifiers.
\newblock {\em arXiv preprint arXiv:2106.06946}, 2021.

\bibitem{lakshminarayanan2017simple}
Balaji Lakshminarayanan, Alexander Pritzel, and Charles Blundell.
\newblock Simple and scalable predictive uncertainty estimation using deep
  ensembles.
\newblock {\em Advances in neural information processing systems}, 30, 2017.

\bibitem{gal2016dropout}
Yarin Gal and Zoubin Ghahramani.
\newblock Dropout as a bayesian approximation: Representing model uncertainty
  in deep learning.
\newblock In {\em international conference on machine learning}, pages
  1050--1059. PMLR, 2016.

\bibitem{durasov2021masksembles}
Nikita Durasov, Timur Bagautdinov, Pierre Baque, and Pascal Fua.
\newblock Masksembles for uncertainty estimation.
\newblock In {\em Proceedings of the IEEE/CVF Conference on Computer Vision and
  Pattern Recognition}, pages 13539--13548, 2021.

\bibitem{dau2019ucr}
Hoang~Anh Dau, Anthony Bagnall, Kaveh Kamgar, Chin-Chia~Michael Yeh, Yan Zhu,
  Shaghayegh Gharghabi, Chotirat~Ann Ratanamahatana, and Eamonn Keogh.
\newblock The ucr time series archive.
\newblock {\em IEEE/CAA Journal of Automatica Sinica}, 6(6):1293--1305, 2019.

\bibitem{athalye2018obfuscated}
Anish Athalye, Nicholas Carlini, and David Wagner.
\newblock Obfuscated gradients give a false sense of security: Circumventing
  defenses to adversarial examples.
\newblock In {\em International conference on machine learning}, pages
  274--283. PMLR, 2018.

\bibitem{papernot2016distillation}
Nicolas Papernot, Patrick McDaniel, Xi~Wu, Somesh Jha, and Ananthram Swami.
\newblock Distillation as a defense to adversarial perturbations against deep
  neural networks.
\newblock In {\em 2016 IEEE symposium on security and privacy (SP)}, pages
  582--597. IEEE, 2016.

\bibitem{yin2022defending}
Sheng-lin Yin, Xing-lan Zhang, and Li-yu Zuo.
\newblock Defending against adversarial attacks using spherical sampling-based
  variational auto-encoder.
\newblock {\em Neurocomputing}, 478:1--10, 2022.

\bibitem{nie2022diffusion}
Weili Nie, Brandon Guo, Yujia Huang, Chaowei Xiao, Arash Vahdat, and Anima
  Anandkumar.
\newblock Diffusion models for adversarial purification.
\newblock {\em arXiv preprint arXiv:2205.07460}, 2022.

\bibitem{madry2017towards}
Aleksander Madry, Aleksandar Makelov, Ludwig Schmidt, Dimitris Tsipras, and
  Adrian Vladu.
\newblock Towards deep learning models resistant to adversarial attacks.
\newblock {\em arXiv preprint arXiv:1706.06083}, 2017.

\bibitem{van2014renyi}
Tim Van~Erven and Peter Harremos.
\newblock R{\'e}nyi divergence and kullback-leibler divergence.
\newblock {\em IEEE Transactions on Information Theory}, 60(7):3797--3820,
  2014.

\bibitem{qin2021dynamic}
Ruoxi Qin, Linyuan Wang, Xingyuan Chen, Xuehui Du, and Bin Yan.
\newblock Dynamic defense approach for adversarial robustness in deep neural
  networks via stochastic ensemble smoothed model.
\newblock {\em arXiv preprint arXiv:2105.02803}, 2021.

\bibitem{liu2020enhancing}
Chizhou Liu, Yunzhen Feng, Ranran Wang, and Bin Dong.
\newblock Enhancing certified robustness via smoothed weighted ensembling.
\newblock {\em arXiv preprint arXiv:2005.09363}, 2020.

\bibitem{karim2017lstm}
Fazle Karim, Somshubra Majumdar, Houshang Darabi, and Shun Chen.
\newblock Lstm fully convolutional networks for time series classification.
\newblock {\em IEEE access}, 6:1662--1669, 2017.

\bibitem{chen2021multi}
Wei Chen and Ke~Shi.
\newblock Multi-scale attention convolutional neural network for time series
  classification.
\newblock {\em Neural Networks}, 136:126--140, 2021.

\bibitem{tsipras2018robustness}
Dimitris Tsipras, Shibani Santurkar, Logan Engstrom, Alexander Turner, and
  Aleksander Madry.
\newblock Robustness may be at odds with accuracy.
\newblock {\em arXiv preprint arXiv:1805.12152}, 2018.

\bibitem{anderson2022certified}
Brendon~G Anderson and Somayeh Sojoudi.
\newblock Certified robustness via locally biased randomized smoothing.
\newblock In {\em Learning for Dynamics and Control Conference}, pages
  207--220. PMLR, 2022.

\end{thebibliography}
\end{document}